\documentclass{article}
\PassOptionsToPackage{numbers, compress}{natbib}

\usepackage{graphicx}
\usepackage[creativeai,final]{neurips_2025}
\usepackage{algorithm}
\usepackage{algpseudocode}
\usepackage{hyperref}
\usepackage{makecell}

\usepackage{amsmath}
\usepackage[utf8]{inputenc} 
\usepackage[T1]{fontenc}    
\usepackage{hyperref}       
\usepackage{url}            
\usepackage{booktabs}       
\usepackage{amsfonts}       
\usepackage{nicefrac}       
\usepackage{microtype}      
\usepackage{xcolor}         

 \title{Text to Robotic Assembly of Multi Component Objects using 3D Generative AI and Vision Language Models}

\author{
Alexander Htet Kyaw\\
Massachusetts Institute of Technology (MIT)\\
\texttt{\small alexkyaw@mit.edu} \\
\And
Richa Gupta \\
MIT\\
\texttt{\small richag@mit.edu} \\
\And
Dhruv Shah \\
Google DeepMind \\
\texttt{\small dhruvshah@google.com} \\
\And
Anoop Sinha \\
Google, Paradigms of Intelligence \\
\texttt{\small anoopsinha@google.com} \\
\And
Kory Mathewson \\
Google DeepMind \\
\texttt{\small korymath@google.com} \\
\And
Stefanie Pender \\
Autodesk Research \\
\texttt{\small stefanie.pender@autodesk.com} \\
\And
Sachin Chitta \\
Autodesk Research \\
\texttt{\small sachin.chitta@autodesk.com} \\
\And
Yotto Koga \\
Autodesk Research \\
\texttt{\small yotto.koga@autodesk.com} \\
\AND
Faez Ahmed \\
MIT Mechanical Engineering \\
\texttt{\small faez@mit.edu} \\
\And
Lawrence Sass \\
MIT Architecture \\
\texttt{\small lsass@mit.edu} \\
\And
Randall Davis \\
MIT CSAIL \\
\texttt{\small davis@csail.mit.edu} \\
}

\begin{document}

\maketitle

\begin{abstract}
Advances in 3D generative AI have enabled the creation of physical objects from text prompts, but challenges remain in creating objects involving multiple component types. We present a pipeline that integrates 3D generative AI with vision-language models (VLMs) to enable the robotic assembly of multi-component objects from natural language. Our method leverages VLMs for zero-shot, multi-modal reasoning about geometry and functionality to decompose AI-generated meshes into multi-component 3D models using predefined structural and panel components. We demonstrate that a VLM is capable of determining which mesh regions need panel components in addition to structural components, based on object functionality. Evaluation across test objects shows that users preferred the VLM-generated assignments 90.6\% of the time, compared to 59.4\% for rule-based and 2.5\% for random assignment. Lastly, the system allows users to refine component assignments through conversational feedback, enabling greater human control and agency in making physical objects with generative AI and robotics.


\end{abstract}

\section{Introduction}
\label{introduction}
Recent developments in 3D Generative AI have enabled users to create a wide variety of geometries from  natural language input \cite{li_generative_2024, gao_get3d_2022, xu_instantmesh_2024}. Extending this capability to make physical objects from a text prompt could empower users who don't have technical expertise in complex 3D design software or manufacturing processes. While previous research has explored physical making with 3D generative AI, challenges remain in creating objects composed of multiple component types with different functionalities \cite{edwards_sketch2prototype_2024, faruqi_style2fab_2023, danry_organs_2023}. Robotic assembly offers a potential approach to create physical objects from multiple component types, while supporting modularity, reuse and part-level editing \cite{jenett_materialrobot_2019, kyaw_speech_2025, kyaw_making_2025}. However, most 3D generative AI models produces monolithic meshes that lack the component-level representation required for robotic assembly path planning and assembly sequencing generation.

\begin{figure} [h]
    \centering
    \includegraphics[width=1\linewidth]{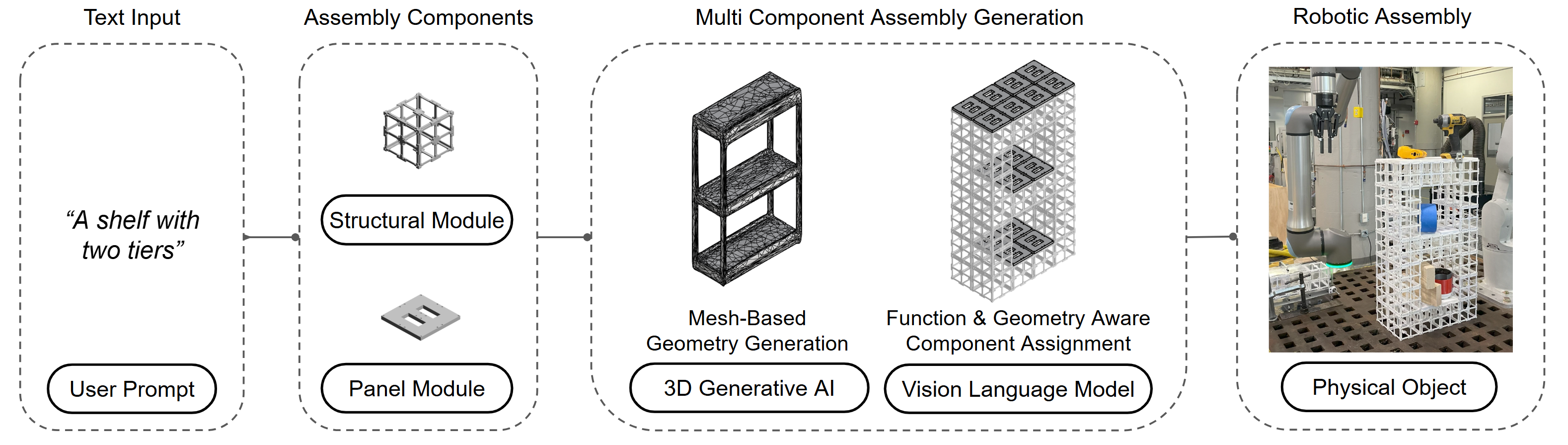}
    \caption {From text input to multi-component robotic assembly using predetermined components}
    \label{fig:teaser}
\end{figure}

Decomposing AI-generated meshes into predefined components is challenging becuase assignments can depend on the geometry and functionality of both the object and its parts. In this study, we use two component types: structural and panel components (Figure \ref{fig:teaser}). While structural components are assigned to the object’s base geometry to ensure stability, the main focus is on determining the placement of panel components based on the object’s geometry and intended function. For example, a stool may require horizontal panels on the seat to create a flat surface for sitting, while a lamp may need panels on its lampshade frame to diffuse light. In addition to geometry and functionality, panel assignments can also vary according to user preferences, highlighting the need for user feedback to support human-AI co-creation for physical making. To address these challenges and enable text-driven, multi-component robotic assembly with generative AI, we present: 

- A function and geometry-aware approach that uses VLM-based multimodal reasoning to assign panel components and decompose AI-generated meshes into multi-component 3D models.

- A conversational feedback workflow using VLMs to enable users to adjust component assignments and provide human control in the AI-driven robotic assembly process without task-specific training.

- An end-to-end framework connecting natural language input, 3D generative AI, VLM, and robotic assembly with predefined components, to create multi-component physical objects from text prompts. 

\section{Related Work}
\textbf{3D Generative AI for Prompt-Based Design and Fabrication.}
Recent advances in text-to-3D models such as DreamFusion \cite{poole_dreamfusion_2022-1}, Get3D \cite{gao_get3d_2022}, and Latte3D \cite{xie_latte3d_2024} have enabled users to generate a wide array of geometries from natural language prompts. To physically realize these models, prior work has mainly explored 3D printing pipelines \cite{danry_organs_2023, westphal_generative_2024}. 
Systems such as Sketch2Prototype transform sketches into printable designs for additive manufacturing \cite{edwards_sketch2prototype_2024}. Style2Fab extends this by allowing users to modify and customize the stylistic attributes of 3D models while preserving printability \cite{faruqi_style2fab_2023}. However, these approaches are for 3D printing with generative AI output that lacks the component-level representation necessary for robotic assembly.

\textbf{Part Aware Generation of 3D Objects.}
Part-aware generative models have enabled structured 3D shape synthesis through component-level reasoning. For example, PartGen \cite{chen_partgen_2024} develops a diffusion-based pipeline that reconstructs semantically meaningful parts from text inputs. StructureNet \cite{mo_structurenet_2019} introduces graph-based part hierarchies geometry synthesis.  ShapeAssembly \cite{jones_shapeassembly_2020} generates objects using a programmatic part layout optimized for visual coherence. While these contributions advance part-level generative modeling, they mostly aim to reconstruct coherent shapes or to support part-level geometry editing. In contrast, our goal is to leverage part-level reasoning to inform the assembly geometry from a predefined set of component type.

\textbf{Component Segmentation of 3D Objects.}
Previous approaches to decomposing 3D shapes into components have relied on supervised methods such as hierarchical recursive networks, point-based segmentation, and semantic graphs~\citep{yu_partnet_2019, charles_pointnet_2017, yi_syncspeccnn_2017}. More recent unsupervised and generative methods, such as Neural Parts~\citep{paschalidou_neural_2021}, auto-decoder frameworks for 3D diffusion models~\citep{ntavelis_autodecoding_2023}, and joint-aware techniques~\citep{li_category-level_2024}, address earlier limitations by introducing flexibility and connection constraints for multi-part reconstruction. However, while these methods improve geometric decomposition, they do not consider robotic assembly or the functional roles of components.

\textbf{Vision-Language Models for Prompt Based  Robotic Assembly.} VLMs have previously been used to ground natural language instructions for assembly tasks. For example, CLIPort \cite{shridhar_cliport_2021} couples CLIP-based perception for pick-and-place tasks. SayCan \cite{ahn_as_2022} integrates language models with affordance scoring to execute multi-step instructions. StructDiffusion \cite{liu_structdiffusion_2023} use diffusion and transformer architectures with multimodal input for compositional tasks. VLMs have also been explored for assembly sequence generation. Zhu et al. \cite{zhu_multi-level_2024} introduces a seq2seq transformer that infers part assembly sequences. Neural Assembler \cite{yan_neural_2025} convert multi-view imagery of block-based models into step-wise assembly plans. These systems demonstrate the potential of VLMs from robotic manipulation to assembly sequence generation. In this paper, we extend the use of VLMs for generating assembly geometry by assigning component types based on object functionality.


\section{Methods}

\textbf{Text‑to‑Mesh Generation.} We present a pipeline using VLM and generative AI to translate natural language inputs into multi-component 3D models for robotic assembly. 
The system begins with a user prompt, which is used to generate a mesh using Autodesk’s 3D generative AI model \cite{malekshan_project_nodate}.

\textbf{Mesh Discretization.} We define two classes of assembly components: (1) structural components, which form the primary load-bearing frame, and (2) panel components, which attach to the structural frame to provide functional surfaces. To create primary load-bearing frame, the AI-generated mesh is discretized into structural components. This is done by voxelizing the mesh using a fixed grid resolution based on the dimensions of the structural components. The placement of panel components depend on the functionality and geometry of the object. Attaching panels indiscriminately can add unnecessary weight and waste components. To address this, we developed a VLM task to selectively assign  faces that should have panel components based on the functionality of the object.

\textbf{VLM for Function Aware Part Selection.}  The VLM is tasked to understand the object’s functionality and geometry to identify the parts requiring a predefined component. In this study we use Google’s Gemini 2.5 pro model. The VLM gives a response based on the three inputs: the description of the object from the user (to understand the object’s functionality), an axonometric image of the AI generated mesh (to understand the object’s geometry), and the component type - in this case the type is panel component (to understand the component’s functionality). Conditioned on these inputs, the VLM is tasked to reason over both functionality and geometry to determine which parts require the component. For example, given the prompt “I want a chair” and the image of the AI‑generated mesh, the VLM returns for panel component “Parts = seat, backrest” as shown in Figure 2. 

System Prompt: You are an assistant that selects the functional parts of an object that require a specified component type. Use: (1) the description of the object, (2) an axonometric image of the AI-generated mesh, and (3) the component type. Select the minimal set of parts that fulfill the object's functionality. Output only the part names as specified, no explanations.

Query: Given an image of \{user text prompt\}, identify which parts of the object should have panel component based on the object's intended functionality. Select only the minimal number of distinct parts required. Output format: Parts = [] 

\begin{figure}[h]
    \centering
    \includegraphics[width=1\linewidth]{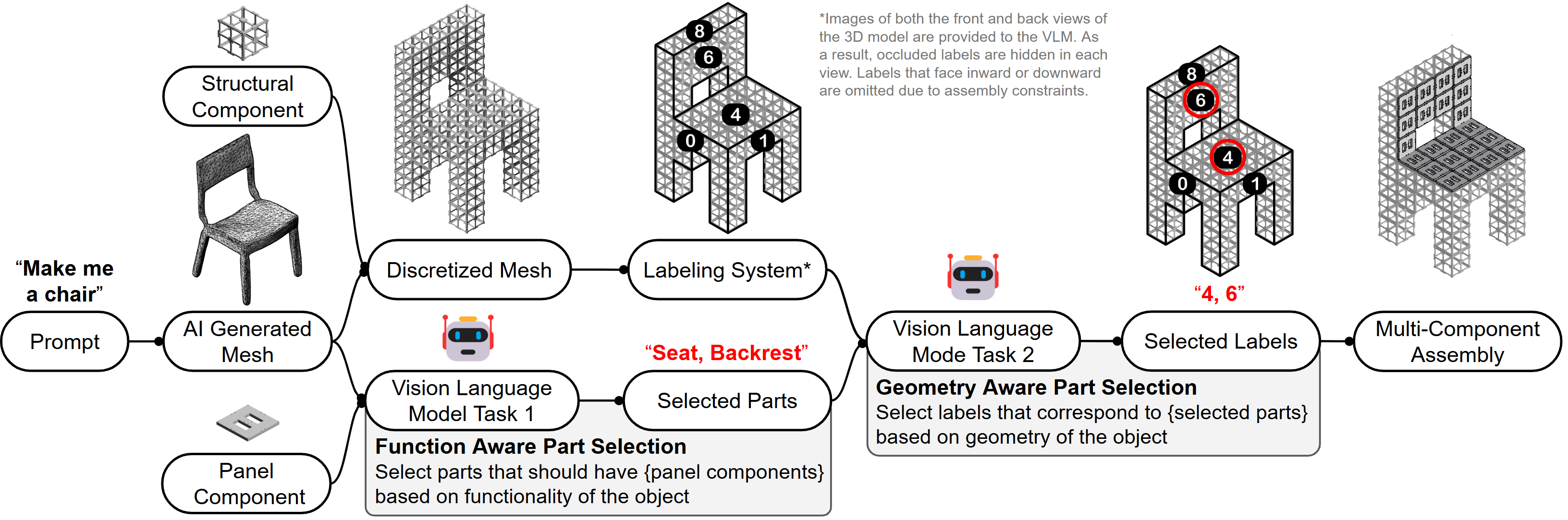}
    \caption{System Pipeline: Vision Language Model for Function and Geometry Aware Part Selection}
    \label{fig:system}
\end{figure}

\begin{figure} [h]
    \centering
    \includegraphics[width=1\linewidth]{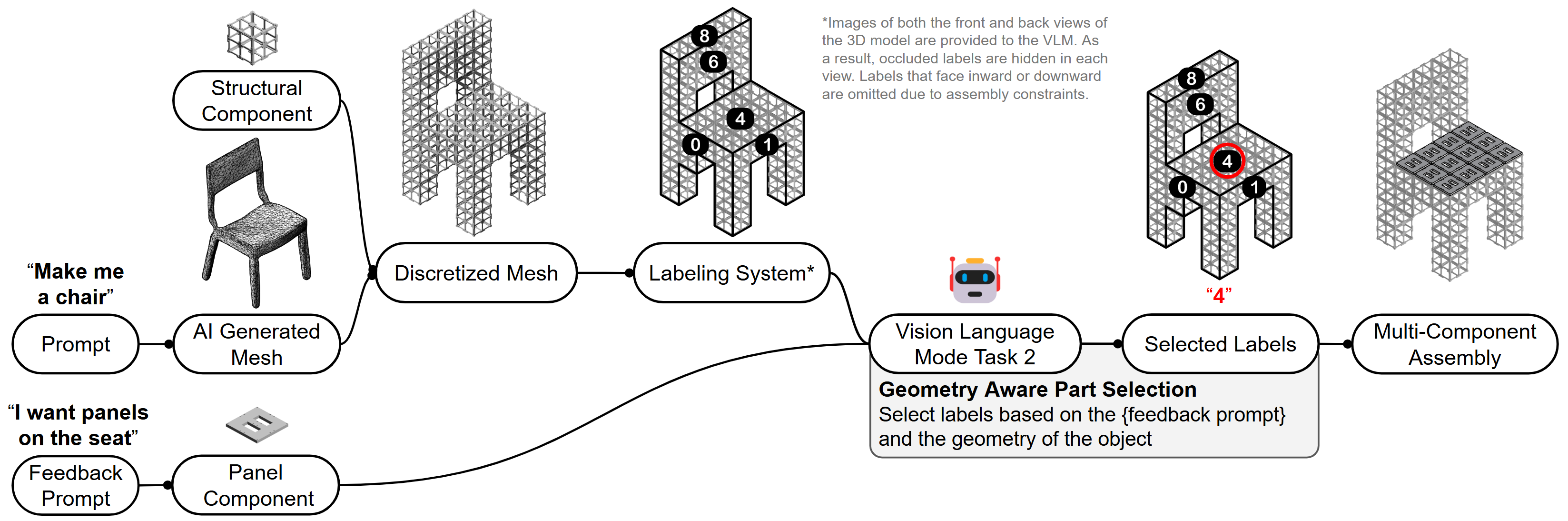}
    \caption {User Alignment: Integrating human feedback with geometry-aware VLM part assignment}
    \label{fig:user}
\end{figure}

\textbf{VLM for Geometry Aware Part Selection.}
The previous VLM task provides a list of parts within the object that should have a panel component. However, the robot must also know where those parts are physically located within the 3S assembly. To map these parts back to the AI‑generated geometry, we merge coplanar faces in the discretized mesh and assign each mesh face a unique integer label for the VLM to reference. Labels for inward‑facing vertical and downward‑facing horizontal faces are omitted, as they might not be reachable by the robot arm. This prevents the model from assigning panels to places the robot can’t access. For the second VLM task, the model receives three inputs: the description of the object from the user, an axonometric rendering of the labeled mesh faces showing both sides of the geometry, and the part list from the first VLM task. Conditioned on these inputs, the VLM matches each part in the list to the corresponding face label. For the example, given the list of parts from the first VLM task “Parts = seat, backrest” and the image of the labeled mesh, the VLM returns “Labels = 4, 6” (Figure~\ref{fig:user}).  The panels are mapped to the labels in the 3D model to create the multi component assembly.

System Prompt: You are an assistant that maps functional parts of an object to their face labels in a labeled axonometric mesh. Use: (1) the description of the object, (2) an image of a labeled mesh, and (3) a list of parts. Select the minimal set of labels that correspond to the listed parts. Output only the label numbers, no explanations.

Query: Given a labeled image of \{prompt\}, select the label numbers that exactly correspond to the following parts: \{parts\}. Select only the minimal set of labels needed.  
Output format: Labels = []

\textbf{VLM for Human in the Loop Conversational User Alignment.}
After the initial component assignment, users can provide feedback to refine or override the VLM generated results. Because human preferences can vary, the system leverages user input instead of relying solely on the VLM. In this task, the user feedback serves as additional context to match the user's intent. The VLM processes three inputs: the description of the object, an axonometric rendering of the labeled mesh faces showing both sides of the geometry, and the user feedback (Figure~\ref{fig:user}). Conditioned on these inputs, VLM outputs the labels corresponding to the user’s feedback. The updated labels are mapped to the mesh to regenerate the multi-component assembly. This human‑in‑the‑loop approach leverages conversational feedback to adjust and control assembly outcomes.

System Prompt: You are an assistant that updates component assignments based on user request. Use: (1) the description of the object, (2) the labeled mesh image, (3) the user request. Select the minimal set of labels that fulfill the user request. Output only the label numbers, no explanations.

Query: Given a labeled image of \{user prompt\}, select the label numbers that match the following request: \{user feedback\}. Select only the minimal set of labels needed. Output format: Labels = []

\textbf{Robotic Assembly} Once the multi-component assembly is generated by the VLM, a UR20 robotic arm equipped with Robotiq grippers assemblies the physical geometry. The multi-component assembly from the VLM is exported as two lists: a coordinate list $C = \{(x_i, y_i, z_i, r_{x_i}, r_{y_i}, r_{z_i})\}$, and a component type list $T = \{t_0, t_1\}$. In our implementation, the type $t_i = 0$ denotes a structural component, which is picked from source $s_i = 0$  (a conveyor belt with structural components), while $t_i = 1$ denotes a panel component, which is picked from source $s_i = 1$ (a stack of panel components). The coordinate list is sorted in a bottom-to-top sequence while preserving connectivity to ensure that physically connected components are placed consecutively.

For each component $i$ in $C$, the robot moves to the source coordinate $s_i$ based on the component type $t_i$, picks the component, moves to the component coordinate $c_i$, and place the component (Figure. \ref{fig:robott}. See algorithm \ref{alg:robotic_assembly} in appendix \ref{fig:robott}  

\begin{figure}[h]
    \centering
    \includegraphics[width=1\linewidth]{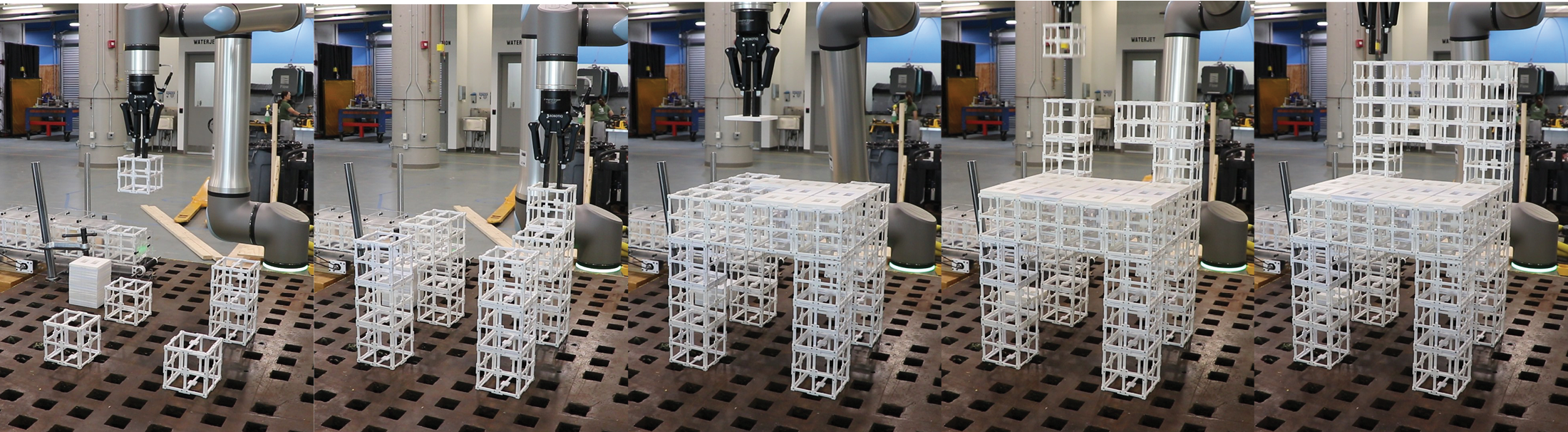}
    \caption{Text to multi-component robotic assembly of the user prompt: "Make me a chair", with the user feedback: "I want panels on the seat".}
    \label{fig:robott}
\end{figure}

\section{Experiments and Results}

\begin{figure}[h]
    \centering
    \includegraphics[width=1\linewidth]{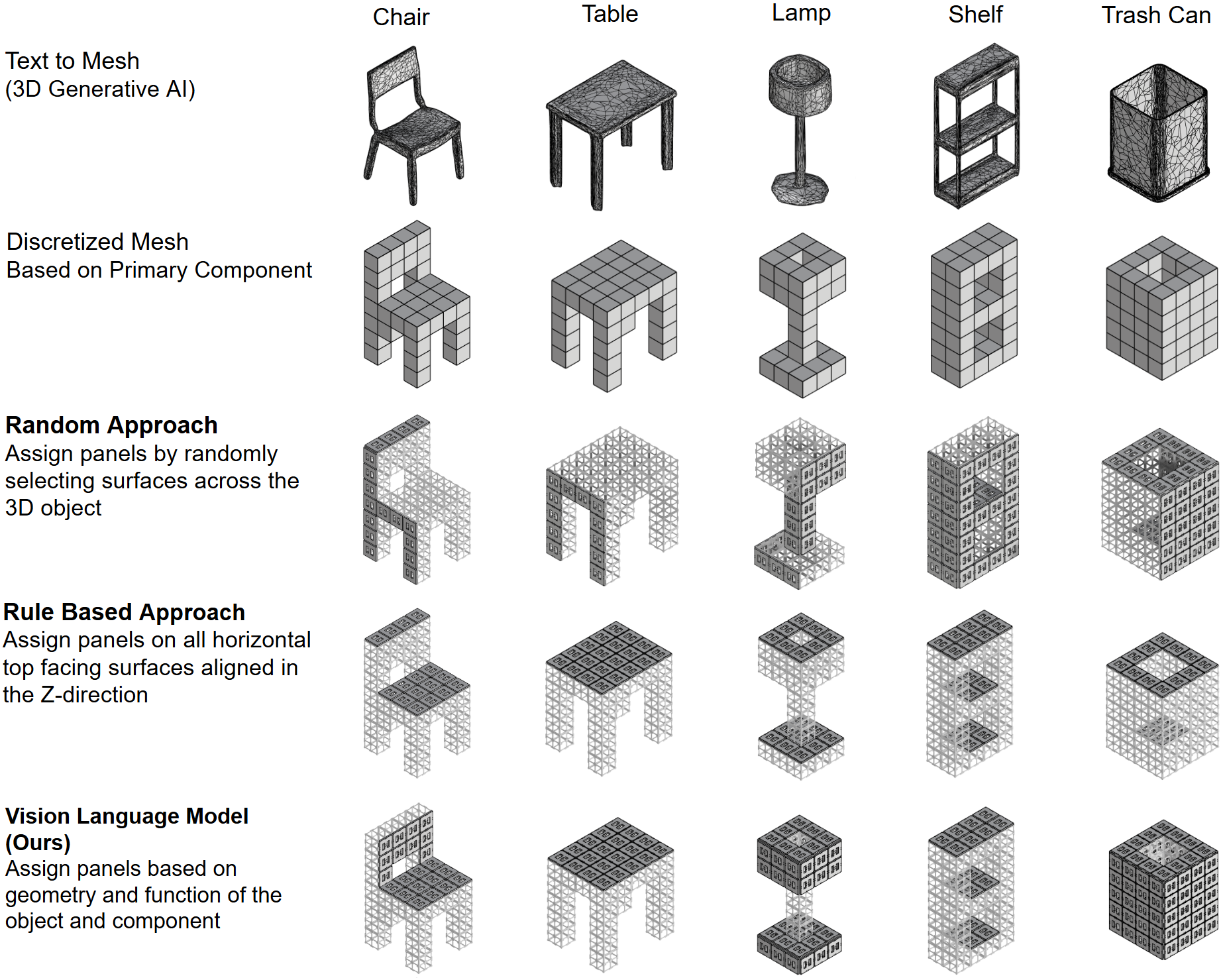}
    \caption{Multi-component assemblies of five different objects created using three different approaches: random, rule-based, and vision-language models}
    \label{fig:matrix}
\end{figure}

\textbf{Experiment Setup.} To assess the effectiveness of the Vision-Language Model (VLM) approach, we conducted a comparative analysis against a \textit{rule-based} approach and a \textit{random} approach. In the rule-based approach, panels were assigned to all upward-facing surfaces, under the assumption that horizontal surfaces are the most likely functional user case for panel components. For the random approach, panels were assigned by randomly selecting labels from the set of mesh faces (Figure \ref{fig:matrix}). 

We recruited thirty two participants to evaluate component assignments on five objects across three approaches, resulting in 480 judgments. Participants were asked to select all alternatives they considered appropriate or acceptable for panel placement based on the object’s function. We assume that multiple valid assignments may exist based on user preference rather than a single ground truth. 

To avoid forced-choice bias, participants were allowed to select multiple options or none for an object. For each object–method pair, we compute the selection rate as:
\[
\text{Rate}_{o,m} = \frac{\# \text{ participants selecting method } m \text{ on object } o}{32} \times 100\%.
\]

\textbf{Results.} 90.6\,\% of users agreed with the VLM generated panel assignments. However, only 59.4\,\% of the user agreed with the rule-based panel assignments. The rule-based approach performs as well as the VLM approach on objects with predominantly horizontal surfaces, such as the table and shelf, but failing on more complex objects like the chair, lamp, and trash can. Unsurprisingly, the random assignment approach performed worst, with a mean selection rate of just 2.5\,\%. 

Additionally, to compare conditions under non-exclusive selection, we applied pairwise McNemar tests, which evaluate responses based on discordant pairs. In this test a higher $\chi^{2}$ indicates a larger imbalance, and thus a stronger preference for one method. The VLM-based approach was chosen significantly more often than both the rule-based approach ($\chi^{2}=38.11$) and the random assignment ($\chi^{2}=137.11$). All pairwise differences remain significant after Bonferroni correction ($^{\ast}p<0.017$). These findings demonstrate that, even without forced-choice constraints, participants preferred VLM-generated placements. (See Appendix~\ref{app:mcnemar} for details.)

\begin{table}[h]
\centering
\caption{Percentage and number of times a user selected a method. Evaluated by 32 participants}

\begin{tabular}{lccccc|c}
\toprule
\textbf{Method} & \textbf{Chair} & \textbf{Table} & \textbf{Lamp} & \textbf{Shelf} & \textbf{Trash Can} & \textbf{Mean} \\
\midrule
VLM (ours)   & 96.9\,\% (31) & 100.0\,\% (32) & 81.3\,\% (26) & 100.0\,\% (32) & 75.0\,\% (24) & 90.6\,\% \\
Rule–based   & 18.8\,\% (6)  & 100.0\,\% (32) & 34.4\,\% (11) & 100.0\,\% (32) & 43.8\,\% (14) & 59.4\,\% \\
Random       &  0.0\,\% (0)  &  0.0\,\% (0)  &  0.0\,\% (0)  &  6.3\,\% (2)   &  6.3\,\% (2)   &  2.5\,\% \\
\bottomrule
\end{tabular}
\label{tab:selection}
\end{table}

\begin{table}[h]
\centering
\caption{Pairwise McNemar tests with Bonferroni correction for the three comparisons ($\alpha = 0.05/3 \approx 0.0167$). Continuity-corrected results are given in Appendix~\ref{app:mcnemar}}
\begin{tabular}{lccc}
\toprule
\textbf{Comparison} & $\chi^{2}$ & \textbf{$p$-value} & \textbf{Conclusion} \\
\midrule
VLM Assignment (ours) \emph{vs.}\ Rule-Based & 38.11 & $<0.001$ & VLM $\gg$ Rule \\
VLM Assignment (ours) \emph{vs.}\ Random    & 137.11 & $<0.001$ & VLM $\gg$ Random \\
Rule-Based \emph{vs.}\ Random               & 88.17 & $<0.001$ & Rule $\gg$ Random \\
\bottomrule
\end{tabular}
\label{tab:mcnemar}
\end{table}


\textbf{Human in the Loop Feedback for VLM Outputs.} Participants also suggested alternative ways of assigning the panel components outside of the three provided methods in the survey. These suggested feedback edits include: applying panels only to the seat of the chair but not the backrest, applying panels to the lampshade but not the base of the lamp, and applying panels only to the bottom two tiers of the shelf. This implies that while the VLM can generate valid outputs, there can be more than one solution. 

As with many design problems, there can be multiple solutions to the same issue, and these can be subject to the user's preferences.
In this case, the AI-generated output serves as a starting point for some users to diverge from. Beyond automation, natural language feedback enables human control in the AI-driven pipeline, accommodating varied user preferences and creative liberties.



\begin{figure} [h]
    \centering
    \includegraphics[width=1\linewidth]{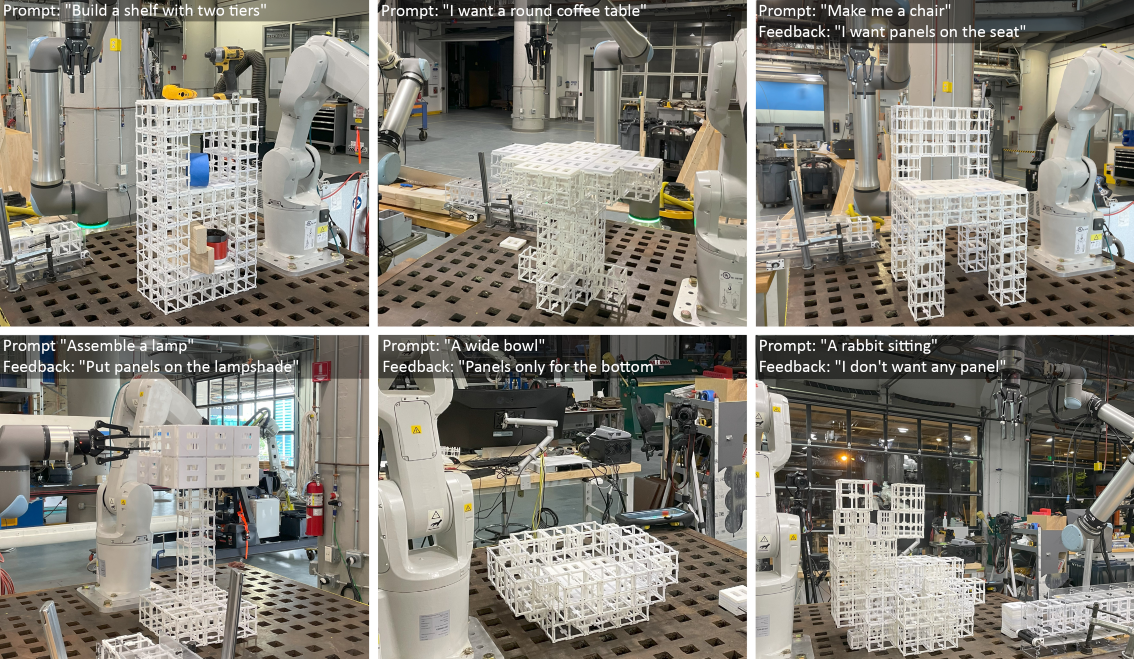}
    \caption{Robotic assembly of select user prompts and feedback, \href{https://youtu.be/gw5ClQtKmAc}{https://youtu.be/ZJrsWG7Mw5M} }
    \label{fig:final}
\end{figure}

\textbf{Robotic Assembly of VLM Generated Designs} The end-to-end robotic assembly pipeline was demonstrated using various user prompts. The robot was able to execute feasible grasp configurations and assembly sequences based on VLM-generated outputs and user-refined outputs. During assembly, the robot did not place any inward- or downward-facing panels, confirming that VLM outputs can comply with fabrication constraints when potential violations are preprocessed out of the input image (Figure. \ref{fig:final}). 

\newpage

\section{Discussion and Limitations}
Our work demonstrates that VLMs can decompose AI-generated meshes into predefined components based on object functionality and geometry, using zero-shot multi-modal reasoning without task-specific training. We attribute the relative success of the VLM-based approach to its broad prior knowledge of object functions, geometric reasoning, and worldview acquired through large-scale multi-modal pretraining. Additionally, unlike the rule-based approach, VLMs can ground their reasoning in both the user prompt and the geometry of the object provided in the input image.

One of the constraints of this study is the use of predefined assembly components. The  current implementation is restricted to two component types. While the narrower scope enabled controlled evaluation and physical assembly with the robotic arm, future work can expand the fixed component library to diversify the types of assembly elements \cite{smith_hierarchical_2025}. This includes extending the VLM pipeline to additional functional components, such as hinges and handles, as well as material-specific components, such as wood, plastic, or metal. Currently, the evaluation is limited to common objects and simple prompts. Future studies should explore the framework with complex user prompts or unconventional objects, which may require multi-turn human-AI interaction with larger edit distances.

This work introduces a pipeline that combines 3D generative AI, VLM-based reasoning, and robotic assembly to transform natural language prompts into functional multi-component objects. In addition, the system supports user edits through natural language feedback, bridging automation with human agency.  Together, these capabilities move toward a vision-language-driven workflow for human-AI co-creation and robotic assembly. 

\newpage

\bibliographystyle{plainnat}
\bibliography{references.bib}


\appendix

\section{Technical Appendices and Supplementary Material}
The following sections include additional details such as supplementary figures, full system prompts, robotic assembly implementation, user survey setup, and full calculation of the statistical analysis results on the experiment data

\subsection{Predefined Component types: Structural and Panel}
\label{app:comp}
\begin{figure}[h]
    \centering
    \includegraphics[width=0.5\linewidth]{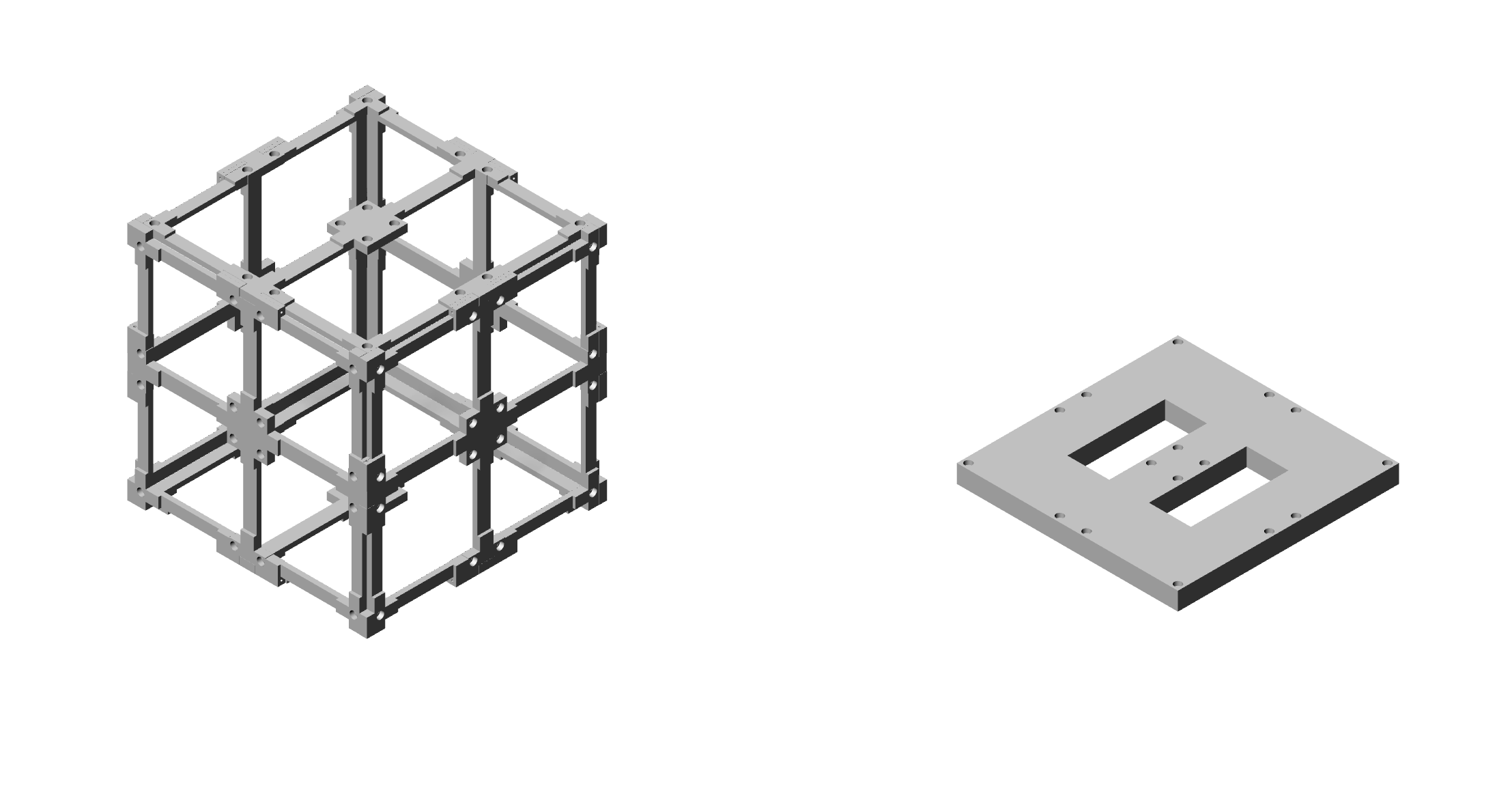}
    \caption{The two predefined component types used in our system.}
    \label{fig:appendix1}
\end{figure}

The structural component is a volumetric cube composed of six faces. It is used for the object’s base geometry to ensure stable robotic assembly. Each side of the structural component has 16 small magnets arranged in a polarity pattern (positive and negative) that ensures proper alignment when the adjacent component connects. The cube’s internal grid structure forms a lattice, making it lightweight while maintaining strength. The top face of the structural component includes a unique feature that allows the robotic arm’s gripper to easily grasp it. The panel component is a flat plane designed to attach to the structural component. To ensure proper connection, the magnet arrangement follows the same polarity pattern as the structural component. Additionally, the panel component includes openings that allow the robotic gripper to securely grasp it during assembly. See figure \ref{fig:appendix1}.

\begin{figure}
    \centering
    \includegraphics[width=0.97\linewidth]{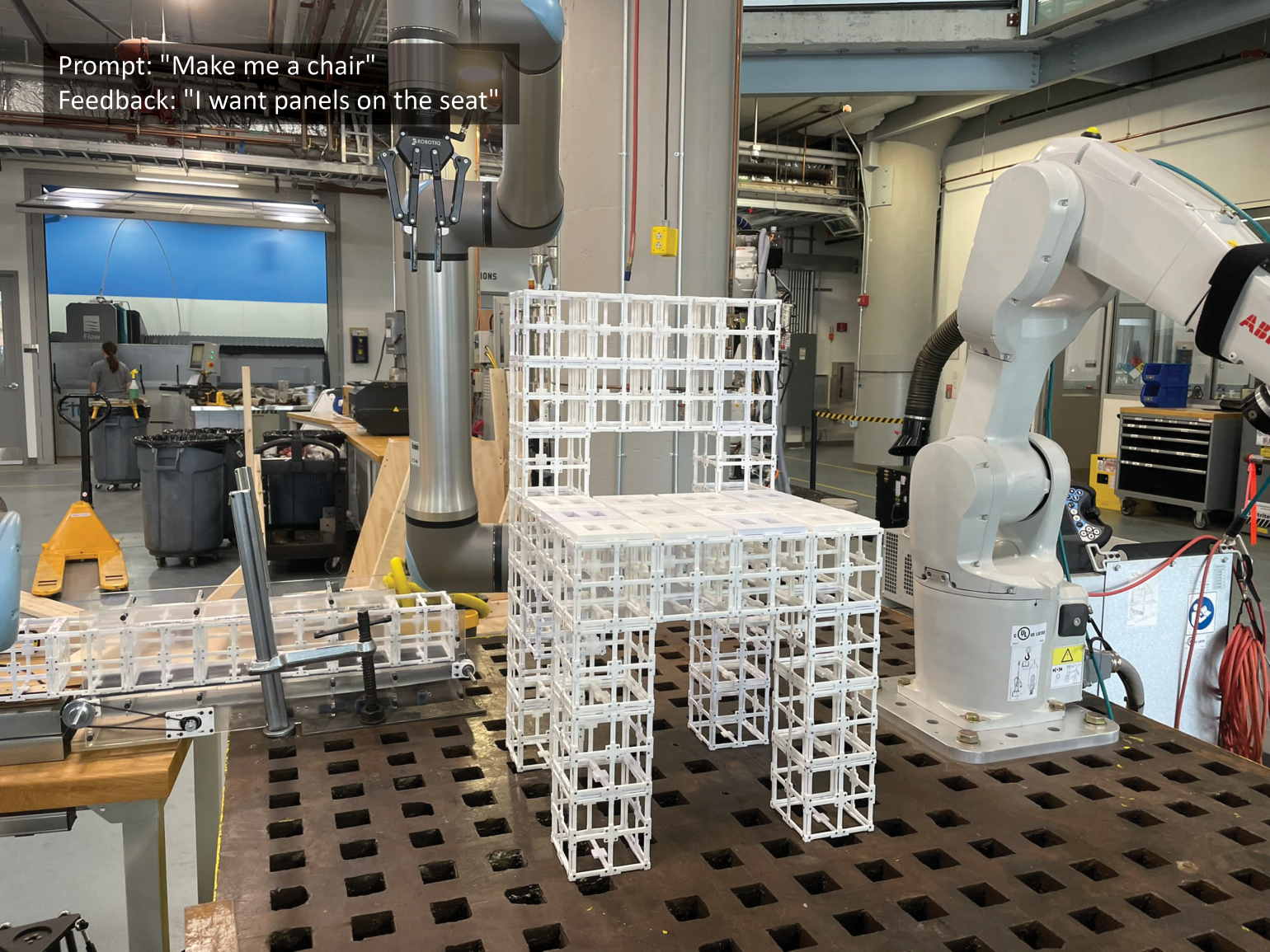}
    \caption{Text to multi-component robotic assembly of the user prompt: "Make me a chair", "I want panels on the seat"}
    \label{fig:placeholder}
\end{figure}

\begin{figure}
    \centering
    \includegraphics[width=0.97\linewidth]{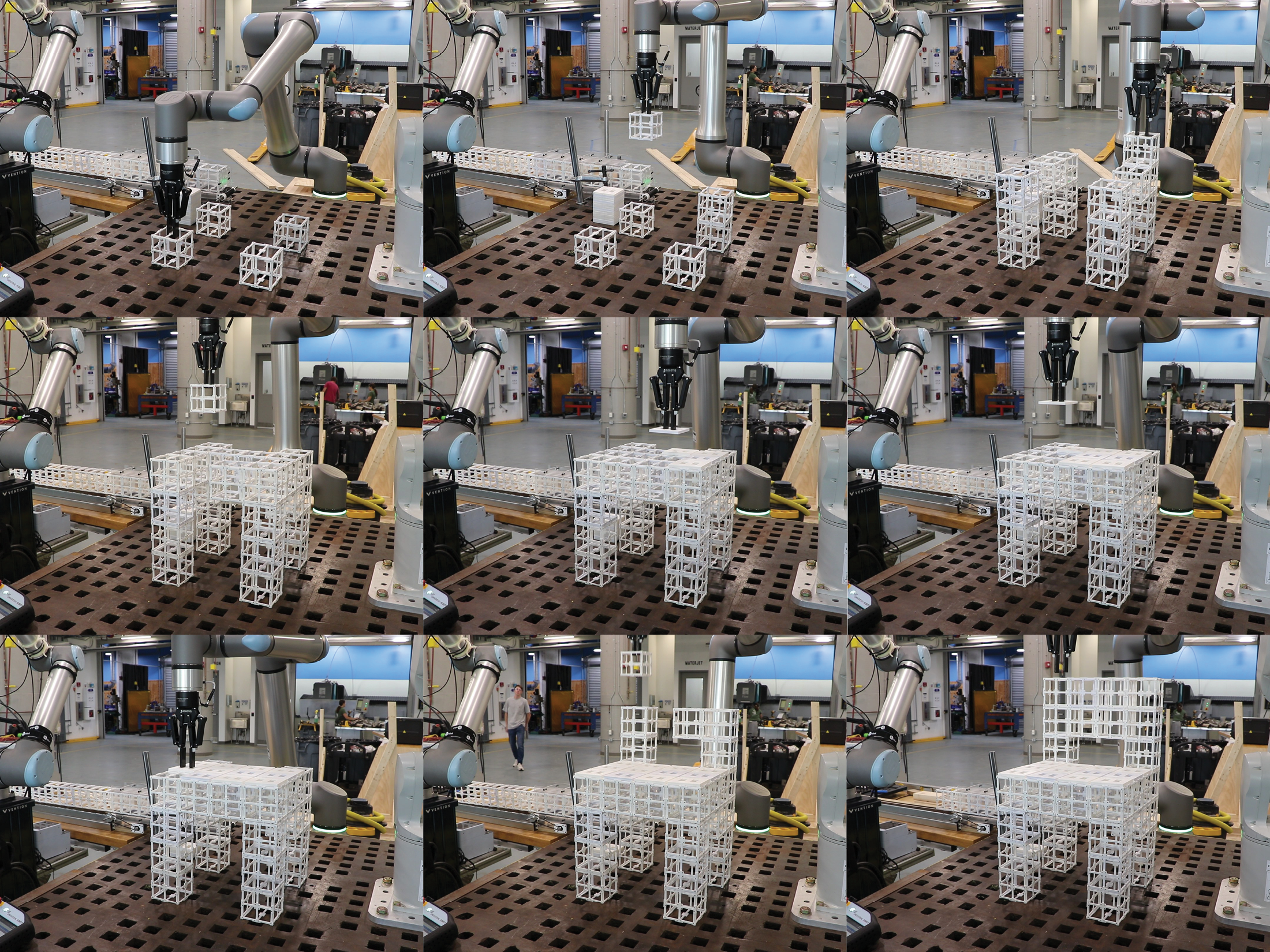}
    \caption{ The robotic assembly process of the prompt: "Make me a chair", "I want panels on the seat"}
    \label{fig:placeholder}
\end{figure}

\begin{figure}
    \centering
    \includegraphics[width=0.97\linewidth]{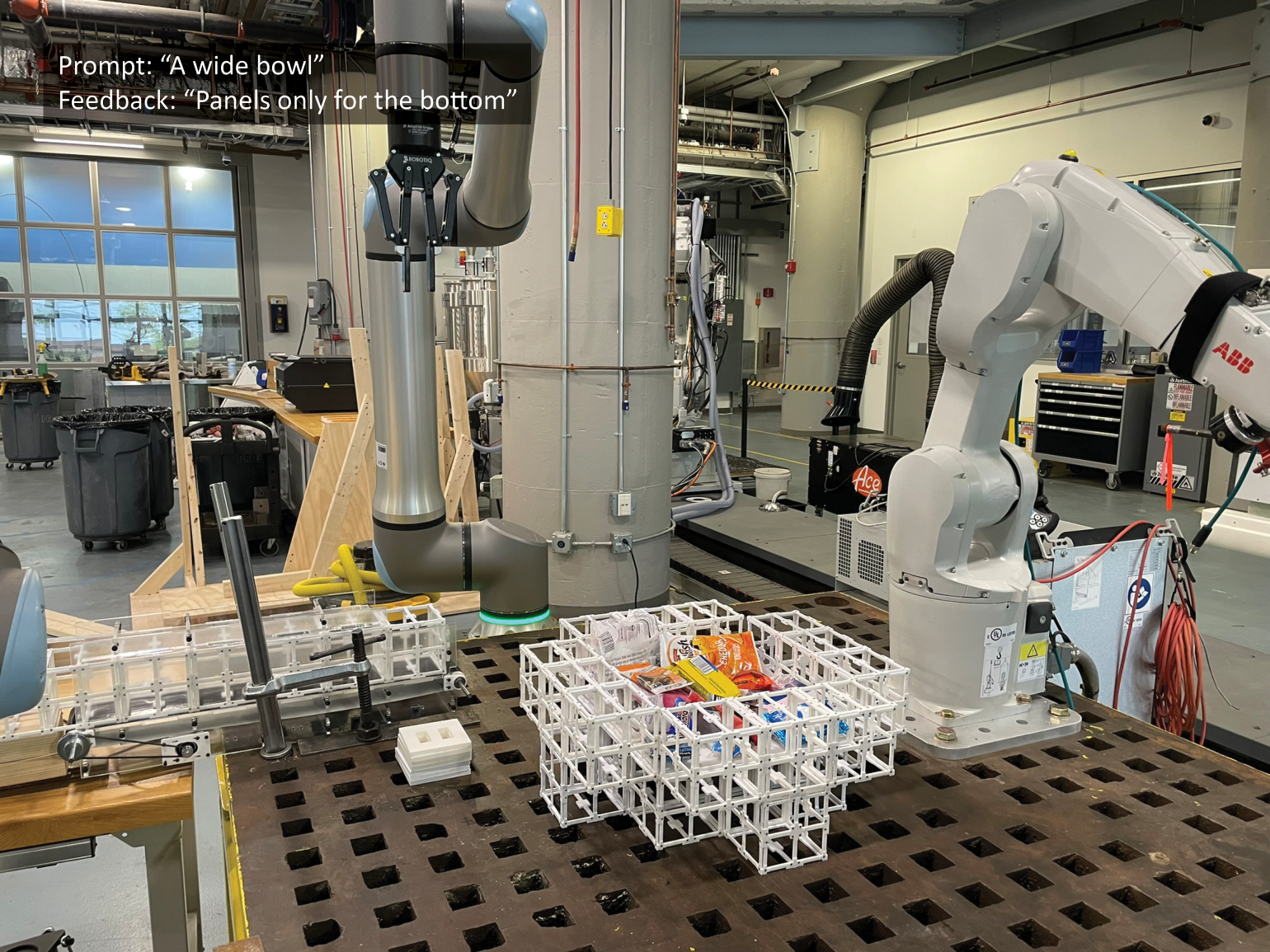}
    \caption{Text to multi-component robotic assembly of the user prompt: "A wide bowl", Panels only for the bottom}
    \label{fig:placeholder}
\end{figure}

\begin{figure}
    \centering
    \includegraphics[width=0.97\linewidth]{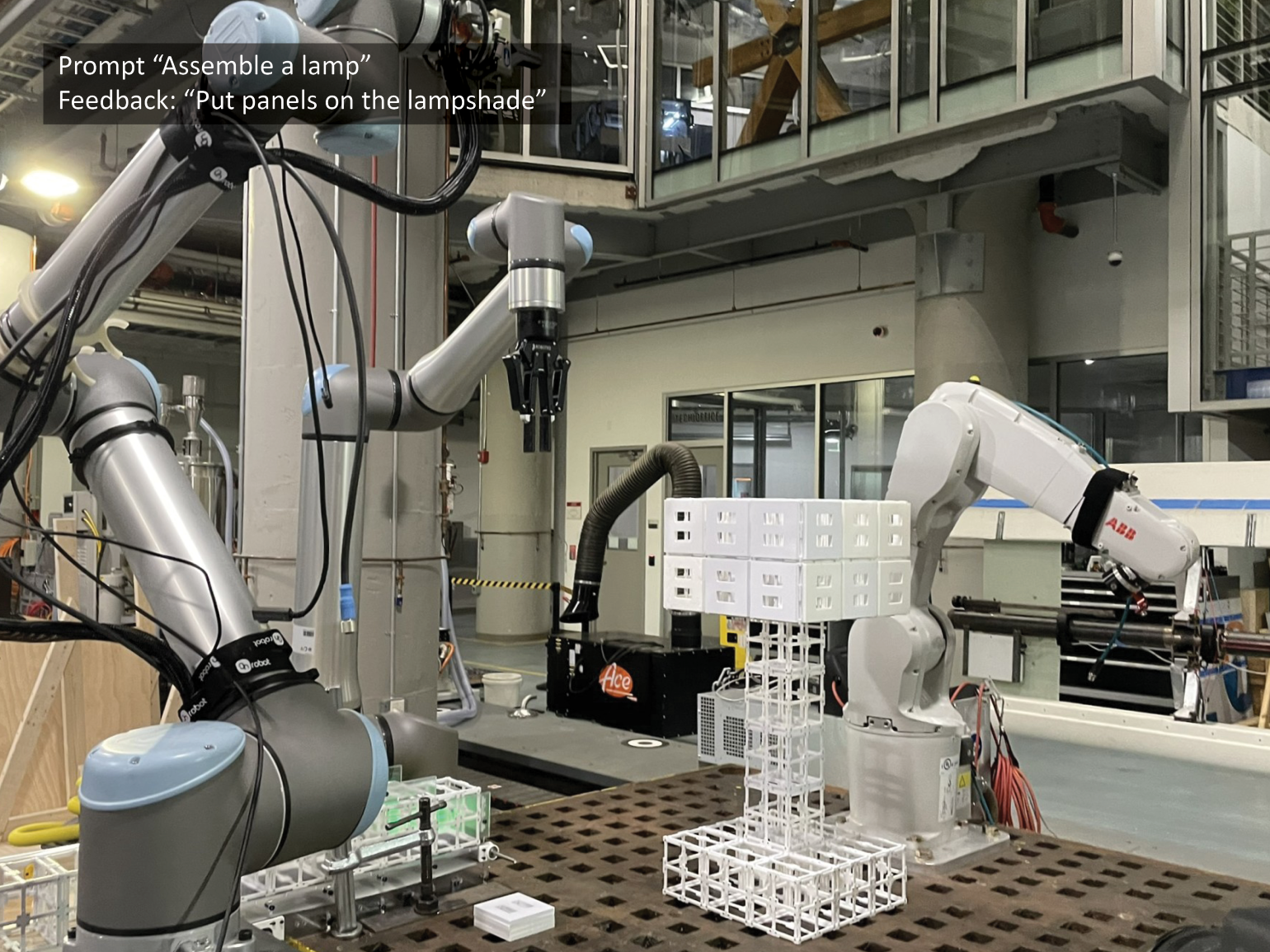}
    \caption{Text to multi-component robotic assembly of the user prompt: "Assemble a lamp", "Put panels on the lampshade"}
    \label{fig:placeholder}
\end{figure}

\begin{figure}
    \centering
    \includegraphics[width=1\linewidth]{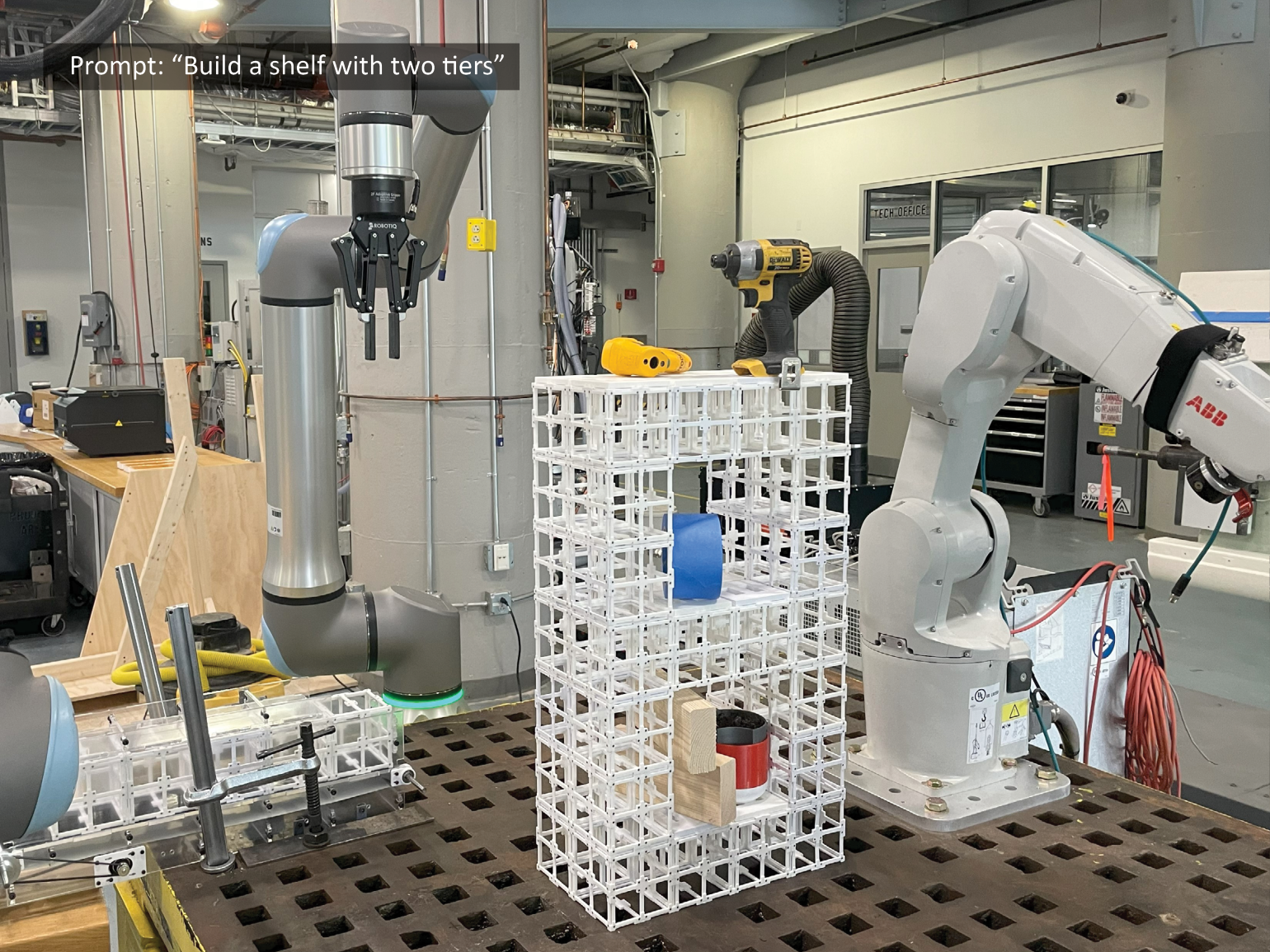}
    \caption{Text to multi-component robotic assembly of the prompt: "Build a shelf with two tiers"}
    \label{fig:placeholder}
\end{figure}

\begin{figure}
    \centering
    \includegraphics[width=1\linewidth]{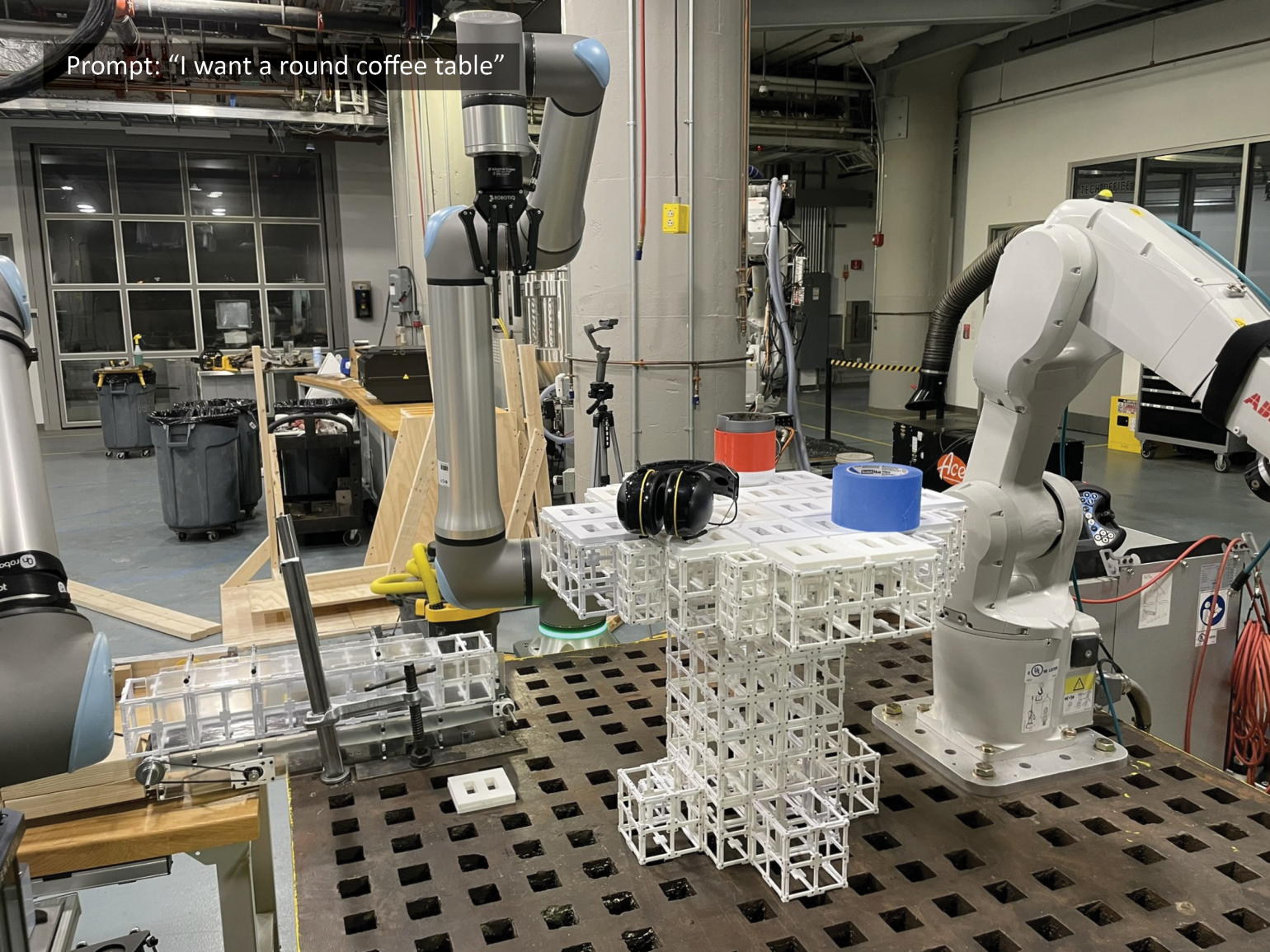}
    \caption{Text to multi-component robotic assembly of the prompt: "I want a round coffee table"}
    \label{fig:placeholder}
\end{figure}

\section{Robotic Assembly Implementation Details}
\label{app:robotic}

The robotic assembly demonstrations are developed using Autodesk's internal robotics research platform integrated with Fusion 360. All experiments were conducted on a standard desktop that can run Python code. 

\begin{algorithm}[H]
\caption{Robotic Assembly Algorithm for Panel Component and Structural Component}
\label{alg:robotic_assembly}
\begin{algorithmic}[1]
\Require
\begin{itemize}
    \item Component coordinates $C = \{(x_i, y_i, z_i, r_{x_i}, r_{y_i}, r_{z_i})\}_{i=1}^n$
    \item Component types $T = \{t_0, t_1\} $ where $t_i \in \{0, 1\}$
    \item Source locations $S_0, S_1$ for each component type
\end{itemize}
\Ensure All component are placed at target positions using pick-and-place operations
\\
\State{\textbf{Initialization}}
\State Initialize robot to home pose $p_{\text{root}}$
\State Open gripper with width $w_{\text{open}}$

\State{\textbf{Iterate through all components in component cooridnates sequence}}
\For{$i = 1$ to $n$}

    \State{\textbf{Determine source location based on component type}}
    \If{$t_i = 0$}  Structural Component
        \State $source \gets S_0$
    \EndIf
    \If{$t_i = 1$}  Panel Component
        \State $source \gets S_1$
    \EndIf
    
    \State{\textbf{Compute pickup and placement poses}}
    \State $pickup \gets (source.x, source.y, source.z)$
    \State $place \gets (x_i, y_i, z_i, r_{x_i}, r_{y_i}, r_{z_i})$

    \State{\textbf{Pick-up sequence}}
    \State Move to $(pickup.x, pickup.y, pickup.z + h_{\text{safe}})$ 
    \State Move to $pickup$ 
    \State Close gripper with force $f_{\text{grab}}$
    \State Move to $(pickup.x, pickup.y, pickup.z + h_{\text{safe}})$ 

    \State{\textbf{Placement sequence}}
    \State Move to $(place.x, place.y, place.z + h_{\text{safe}})$ 
    \State Move to $place$ 
    \State Open gripper with width $w_{\text{release}}$
    \State Move to $(place.x, place.y, place.z + h_{\text{safe}})$

    \State \textbf{Output:} Component $i$ successfully placed
\EndFor
\end{algorithmic}
\end{algorithm}

\section{Additional Details on Experiment and Results}
\label{app:mcnemar}

\textbf{IRB Approval Statement}
This study involving human subjects was reviewed by the Massachusetts Institute of Technology Committee on the Use of Humans as Experimental Subjects (MIT COUHES), which serves as the Institutional Review Board (IRB) for MIT. 

The study received an IRB exemption determination (Exemption Category 3) as it involved minimal risk and no collection of sensitive personal information. All participants were adults who provided informed consent prior to participating. No personally identifying information was collected.

Before beginning the survey, participants were informed that:
\begin{itemize}
    \item They would be shown five objects generated by our AI-based design system.
    \item Their task would be to evaluate panel placements for each object.
    \item They could skip any question they did not wish to answer.
    \item Their responses would remain anonymous and would be used for academic research.
    \item They would not receive monetary or other forms of compensation for participating.
\end{itemize}

\vspace{4\baselineskip}

\textbf{Instructions for Selection}  
\begin{quote}
\textit{For each object, you will see three possible configurations for where panels could be placed. The configurations were generated using three different methods. Select all configurations that you believe have appropriate or acceptable panel placements for the given object and its function. You may select one, more than one, or none at all. Please select according to what you think is functionally appropriate. Please use the check mark on Adobe PDF to select a checkbox. }
\end{quote}

\begin{figure}[H]
    \centering
    \includegraphics[width=0.6\linewidth]{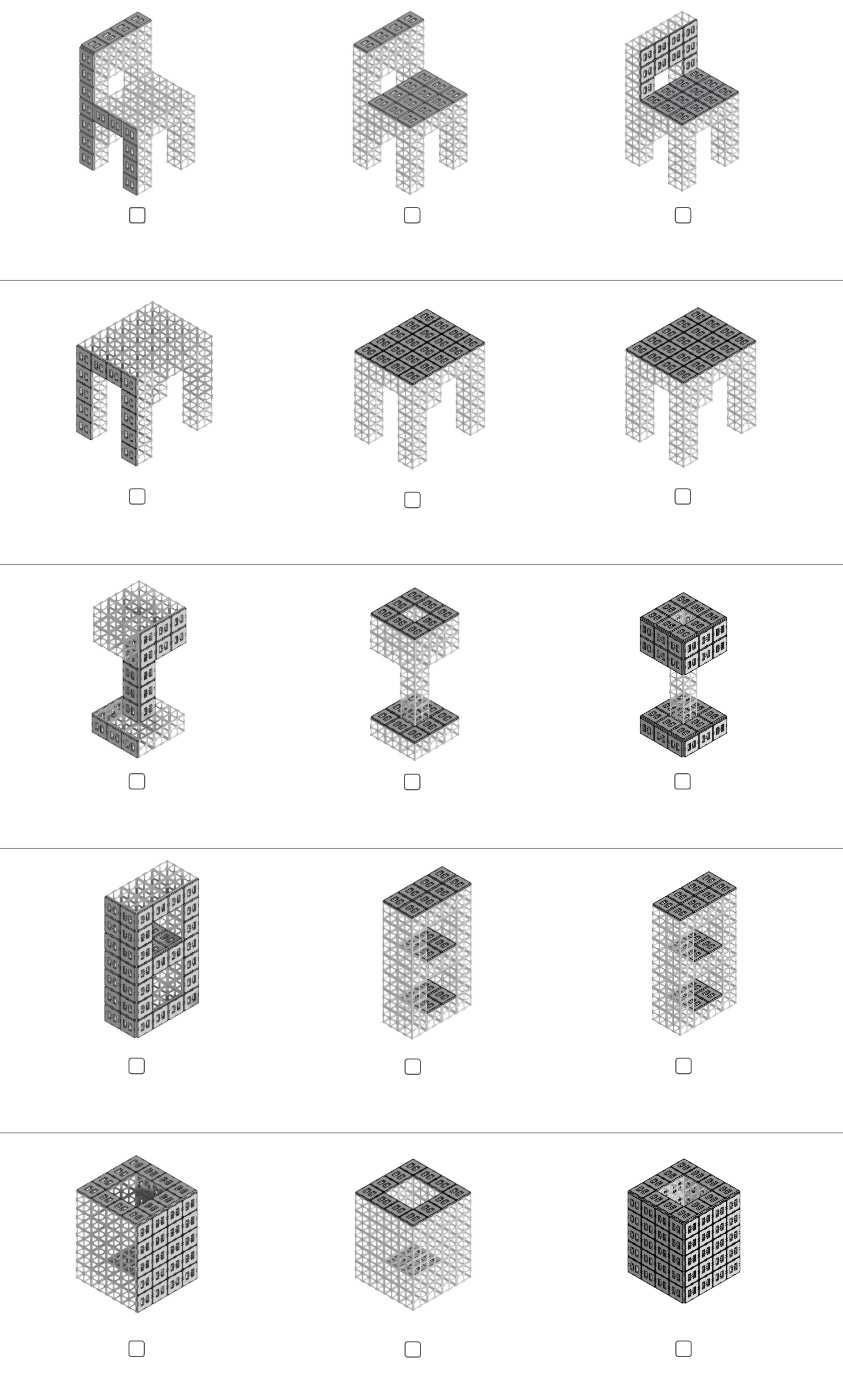}
    \caption{Layout of the survey document with three options of each of the five objects. No labels were provided to avoid bias. A checkbox allowed participants to select each option, and a line was provided after each object for text input.}
    \label{fig:task1_survey}
\end{figure}

\textbf{Instructions for Alternative Assignments}  
\begin{quote}
\textit{For each object, indicate whether you think there is an alternative way to assign panels that is different from the options shown above. If yes, describe how you would assign the panels differently. If no, just write “No.”. Please use Adobe PDF text box for the text entry. } 
\end{quote}

\vspace{4\baselineskip}

\subsection{Derivation of Test Statistics: Pairwise McNemar Tests}

For every \textit{participant–object–method} triple we log a binary outcome  
(1 = “panel placement judged appropriate”, 0 = “not selected”).  
When comparing two methods, A and B, we form a $2\times2$ contingency
table and consider only the \emph{discordant} cells:

\begin{table}[h]
\centering
\caption{Variables for McNemar Calculation.}
\begin{tabular}{lcc}
\toprule
\textbf{Participant’s judgement on that object} & \textbf{Contributes to} \\
\midrule
A = 1,\;B = 0 \:(A chosen,\;B not) & $b$ \\[2pt]
A = 0,\;B = 1 \:(B chosen,\;A not) & $c$ \\[2pt]
A = 1,\;B = 1 \textit{or} A = 0,\;B = 0 &  (concordant; ignored) \\
\bottomrule
\end{tabular}
\label{tab:mcnemar-varaible}
\end{table}

\paragraph{Test statistic.}
McNemar’s null hypothesis is $b=c$. We report the  \emph{uncorrected}
statistic and, in parentheses, the continuity‑corrected version:
\[
\chi^{2}_{\text{uncorr}}=\frac{(b-c)^{2}}{b+c}, \qquad
\chi^{2}_{\text{corr}}=\frac{(|b-c|-1)^{2}}{b+c}.
\]

\begin{table}[h]
\centering
\caption{Discordant counts and McNemar results (
160 trials per comparison for the five objects)}
\begin{tabular}{lcccccc}
\toprule 
\textbf{Comparison} & $b$ & $c$ & $\chi^{2}_{\text{uncorr}}$ & $p$ & $\chi^{2}_{\text{corr}}$ & $p$ \\ 
\midrule
VLM \textit{vs.}\ Rule & 56 & 7 & 38.11 & $<0.001$ & 36.57 & $<0.001$ \\
VLM \textit{vs.}\ Random & 143 & 2 & 137.11& $<0.001$ & 135.17& $<0.001$ \\
Rule \textit{vs.}\ Random & 94 & 2 & 88.17 & $<0.001$ & 86.26 & $<0.001$ \\
\bottomrule
\end{tabular}
\label{tab:mcnemar-work-32}
\end{table}

\noindent
All $p$‑values remain well below the Bonferroni‑adjusted threshold
$\alpha=0.05/3\approx0.0167$; applying the continuity correction does not change any conclusion.

\paragraph{Interpretation.}
A larger $\chi^{2}$ indicates a stronger imbalance between the two discordant
cells. For example, more participants selected one method but not the other.  
The results confirm that VLM‑generated placements were overwhelmingly
preferred over both baselines, even though participants were free to select 
multiple methods for the same object.





\newpage
\section*{NeurIPS Paper Checklist}

\begin{enumerate}

\item {\bf Claims}
    \item[] Question: Do the main claims made in the abstract and introduction accurately reflect the paper's contributions and scope?
    \item[] Answer: \answerYes{} 
    \item[] Justification:
        The introduction lists three concrete contributions, and these same contributions are implemented in the Methods section and validated in Experiments. 

\item {\bf Limitations}
    \item[] Question: Does the paper discuss the limitations of the work performed by the authors?
    \item[] Answer: \answerYes{}
    \item[] Justification: Section 5 discusses several limitations of the proposed work, including the use of a fixed library and the evaluation being limited to simple prompts. The authors also note that future extensions should support a broader range of functional components and multi-turn user interaction.

\item {\bf Theory assumptions and proofs}
    \item[] Question: For each theoretical result, does the paper provide the full set of assumptions and a complete (and correct) proof?
    \item[] Answer:  \answerNA{}.
    \item[] Justification:  The paper does not present any formal theoretical results or mathematical theorems. It focuses on a novel system design and empirical evaluation of VLM-based reasoning for robotic assembly.

\item {\bf Experimental result reproducibility}
    \item[] Question: Does the paper fully disclose all the information needed to reproduce the main experimental results of the paper to the extent that it affects the main claims and/or conclusions of the paper (regardless of whether the code and data are provided or not)?
    \item[] Answer: \answerYes{} 
    \item[] Justification: The paper provides detailed descriptions of the experimental setup, including the user study, VLM prompting tasks, mesh discretization, component assignment logic, and user study protocol. It also includes algorithmic pseudocode for robotic assembly and statistical analysis procedures. 

\item {\bf Open access to data and code}
    \item[] Question: Does the paper provide open access to the data and code, with sufficient instructions to faithfully reproduce the main experimental results, as described in supplemental material?
    \item[] Answer:  \answerNo{} .
    \item[] Justification: The code for Autodesk’s internal robotics platform and Project Bernini is currently under active research and remains closed-source / proprietary. However, the paper offers detailed descriptions of the key methodology and how alternative tools can be used as substitutes for the framework in the Appendix. The paper also provides open access to the VLM prompts and the robotic assembly algorithm in the appendix.

\item {\bf Experimental setting/details}
    \item[] Question: Does the paper specify all the training and test details (e.g., data splits, hyperparameters, how they were chosen, type of optimizer, etc.) necessary to understand the results?
    \item[] Answer: \answerYes{}
    \item[] Justification: The paper clearly specifies all relevant experimental details, including the user study, the design of the VLM prompting tasks, the inputs provided (text prompts and images), the mesh discretization process, the predefined component types, and robotic assembly for the readers to understand and interpret the results.  See Appendix \ref{app:robotic}. The paper does not involve training new models, as it uses a pretrained VLM (Gemini 2.5 Pro) in a zero-shot setting. Therefore, there are no training data splits, hyperparameters, or optimizers to report. 

\item {\bf Experiment statistical significance}
    \item[] Question: Does the paper report error bars suitably and correctly defined or other appropriate information about the statistical significance of the experiments?
    \item[] Answer:  \answerYes{}
    \item[] Justification:  The result section and Table \ref{tab:mcnemar} reports chi-square statistics and Bonferroni-corrected p-values for all pairwise comparisons. The statistical test used (McNemar test) is for the non-exclusive selection setup, and the calculation is explained in detail in Appendix \ref{app:mcnemar}.
    
\item {\bf Experiments compute resources}
    \item[] Question: For each experiment, does the paper provide sufficient information on the computer resources (type of compute workers, memory, time of execution) needed to reproduce the experiments?
    \item[] Answer: \answerYes{} 
    \item[] Justification: : The paper specifies that the VLM experiments were conducted using Google's Gemini 2.5 Pro in a zero-shot setting, which does not require local training or significant compute resources. The robotic assembly experiments were conducted using a Universal Robots UR20 industrial robotic arm, with the process described in detail in Appendix \ref{app:robotic}. All experiments were conducted on a standard desktop that can run Python code, which interfaced with Project Bernini for 3D generation, the VLM for reasoning, and Autodesk’s internal robotics platform for robotic programming.
    
\item {\bf Code of ethics}
    \item[] Question: Does the research conducted in the paper conform, in every respect, with the NeurIPS Code of Ethics \url{https://neurips.cc/public/EthicsGuidelines}?
    \item[] Answer: \answerYes{} 
    \item[] Justification: We have reviewed the NeurIPS Code of Ethics and confirm that the research conducted in this paper complies with all outlined principles.

\item {\bf Broader impacts}
    \item[] Question: Does the paper discuss both potential positive societal impacts and negative societal impacts of the work performed?
    \item[] Answer: \answerYes{}
    \item[] Justification: The paper discusses broader impacts in the Introduction and the Discussion section, highlighting societal implications of the proposed system. The Introduction section notes that the system enables accessible physical making for non-experts through natural language. We also mentioned how robotic assembly could support modular construction and part-level editing for component reuse. The discussion section calls for further study and impact, including handling complex prompts and ways to add human oversight and control in creative workflows with AI and robotic systems.
    
\item {\bf Safeguards}
    \item[] Question: Does the paper describe safeguards that have been put in place for responsible release of data or models that have a high risk for misuse (e.g., pretrained language models, image generators, or scraped datasets)?
    \item[] Answer: \answerNA{}.
    \item[] Justification:  The paper does not release any new pretrained models, large-scale datasets, or generative tools that pose a high risk of misuse. It uses existing closed-source tools (e.g., Gemini 2.5 Pro, Project Bernini) and focuses on system integration and framework design.
  
\item {\bf Licenses for existing assets}
    \item[] Question: Are the creators or original owners of assets (e.g., code, data, models), used in the paper, properly credited and are the license and terms of use explicitly mentioned and properly respected?
    \item[] Answer: \answerYes{}
    \item[] Justification: All third-party assets used in the paper, including Google’s Gemini 2.5 Pro vision-language model and Autodesk’s Project Bernini, are properly credited in the text and citations. These tools are accessed in accordance with their respective terms of use. The paper does not redistribute these assets but uses them as services or platforms within the scope allowed for academic research. No external code or datasets with restrictive licenses are included or released.

\item {\bf New assets}
    \item[] Question: Are new assets introduced in the paper well documented and is the documentation provided alongside the assets?
    \item[] Answer: \answerYes{}
    \item[] Justification:  While the paper does not release new datasets, models, or code, it introduces new assets in the form of figures, videos, and physically assembled objects generated through the proposed pipeline. These assets are documented throughout the paper and the appendix.

\item {\bf Crowdsourcing and research with human subjects}
    \item[] Question: For crowdsourcing experiments and research with human subjects, does the paper include the full text of instructions given to participants and screenshots, if applicable, as well as details about compensation (if any)? 
    \item[] Answer: \answerYes{}
    \item[] Justification:  The paper includes a user study involving 25 participants who evaluated component assignments across five objects. The study design is described in the main paper, including the number of participants, object conditions, and the non-exclusive selection task. Participants were recruited and completed the survey voluntarily, providing informed consent. More details on Appendix \ref{app:mcnemar}.

\item {\bf Institutional review board (IRB) approvals or equivalent for research with human subjects}
    \item[] Question: Does the paper describe potential risks incurred by study participants, whether such risks were disclosed to the subjects, and whether Institutional Review Board (IRB) approvals (or an equivalent approval/review based on the requirements of your country or institution) were obtained?
    \item[] Answer: \answerYes{}
    \item[] Justification: The user study posed minimal risk to participants. We obtained IRB exemption through our institution’s ethics review process, which determined that the study qualifies as exempt human subjects research (Appendix \ref{app:mcnemar}) Participants were informed of the study's purpose, participated voluntarily, and provided informed consent prior to completing the survey. No sensitive personal data was collected.

\item {\bf Declaration of LLM usage}
    \item[] Question: Does the paper describe the usage of LLMs if it is an important, original, or non-standard component of the core methods in this research? Note that if the LLM is used only for writing, editing, or formatting purposes and does not impact the core methodology, scientific rigorousness, or originality of the research, declaration is not required.
    \item[] Answer: \answerYes{} 
    \item[] Justification:  The paper uses a language model specifically, Google’s Gemini 2.5 Pro as a core component of the methodology. The VLM performs zero-shot, multimodal reasoning to assign functional components to AI-generated 3D meshes based on input prompts and object geometry.

\end{enumerate}

\end{document}